\documentclass[runningheads]{llncs}
\usepackage[T1]{fontenc}
\usepackage[hidelinks]{hyperref}
\usepackage[numbers, compress]{natbib}
\usepackage{graphicx}
\usepackage{booktabs}
\usepackage{multirow}
\usepackage{comment}
\usepackage{enumitem}
\usepackage{xcolor}

\usepackage[ruled,vlined]{algorithm2e} 
\usepackage{amsmath} 
\usepackage{booktabs}
\usepackage{siunitx}

\begin{document}

\title{Classifying the Unknown:  In-Context Learning for Open-Vocabulary Text and Symbol Recognition}

\titlerunning{Classifying the Unknown with Rosetta}

\author{Tom Simon\inst{1} \and William Mocaer\inst{1} \and
Pierrick Tranouez\inst{1}
\and Clément Chatelain\inst{2} \and
Thierry Paquet \inst{1}}
\authorrunning{T. SIMON et al.}

\institute{LITIS EA4108, University of Rouen Normandy, France
\email{\{tom.simon,william.mocaer, pierrick.tranouez, thierry.paquet\}@univ-rouen.fr}\\ 
\and LITIS EA4108, INSA of Rouen Normandy, France
\email{clement.chatelain@insa-rouen.fr}
}

\maketitle  
\vspace{-5mm}
\begin{abstract}
We introduce \textbf{Rosetta}, a multimodal model that leverages Multimodal In-Context Learning (MICL) to classify sequences of novel script patterns in documents by leveraging minimal examples, thus eliminating the need for explicit retraining. To enhance contextual learning, we designed a dataset generation process that ensures varying degrees of contextual informativeness, improving the model’s adaptability in leveraging context across different scenarios. A key strength of our method is the use of a Context-Aware Tokenizer (CAT), which enables open-vocabulary classification. This allows the model to classify text and symbol patterns across an unlimited range of classes, extending its classification capabilities beyond the scope of its training alphabet of patterns. As a result, it unlocks applications such as the recognition of new alphabets and languages. Experiments on synthetic datasets demonstrate the potential of Rosetta to successfully classify Out-Of-Distribution visual patterns and diverse sets of alphabets and scripts, including but not limited to Chinese, Greek, Russian, French, Spanish, and Japanese. \footnote{The code and dataset from this study will soon be accessible as open-source on the GitHub repository: \url{https://github.com/TSResearch-hub/Rosetta-ICL-classif}.}


\end{abstract}

\vspace{-5mm}
\section{Introduction}
Text and symbol recognition face significant limitations in classifying Out-of-Distribution (OOD) data, i.e., data not encountered during training. Classifier models, such as Optical Character Recognition (OCR) models, are dependent on script, font, and language, making them ineffective when faced with novel text or symbol patterns, and previously unseen alphabets. These models require fine-tuning to perform on such data. This limitation stems from the conventional training paradigm, in which models encode fixed associations between visual patterns and class labels, limiting their ability to adapt to new patterns with a significant distribution shift. Furthermore, because these models rely on a limited vocabulary of classes, they are inherently constrained to predicting only the classes they were trained on, thereby excluding the possibility of open-vocabulary classification. This limitation is especially critical when dealing with unfamiliar alphabets or languages.

To overcome these limitations, we leverage recent advancements in \textbf{In-Context Learning} (ICL), which exploit context to allow model adaptation at inference-time \cite{gpt3,icl_explanation}. By leveraging the ICL paradigm, a model can process a prompt containing multiple input-output examples, in a few-shot manner, to generalize and generate accurate predictions on new data. Through dynamic contextual adaptation, this approach eliminates the need for additional training, providing a scalable and efficient alternative to traditional fine-tuning methods. While existing ICL-based approaches \cite{flamingo,openflamingo,otter,idefics,idefics2,kosmos,mmicl,emu2,mm1,blip3} primarily integrate large language models (LLMs) for tasks such as image captioning and Visual Question Answering (VQA), where linguistic understanding is crucial, we propose a model specifically optimized for a vision task, entirely independent of language knowledge.

\begin{figure}[h]
  \centering
  \resizebox{\textwidth}{!}{\includegraphics{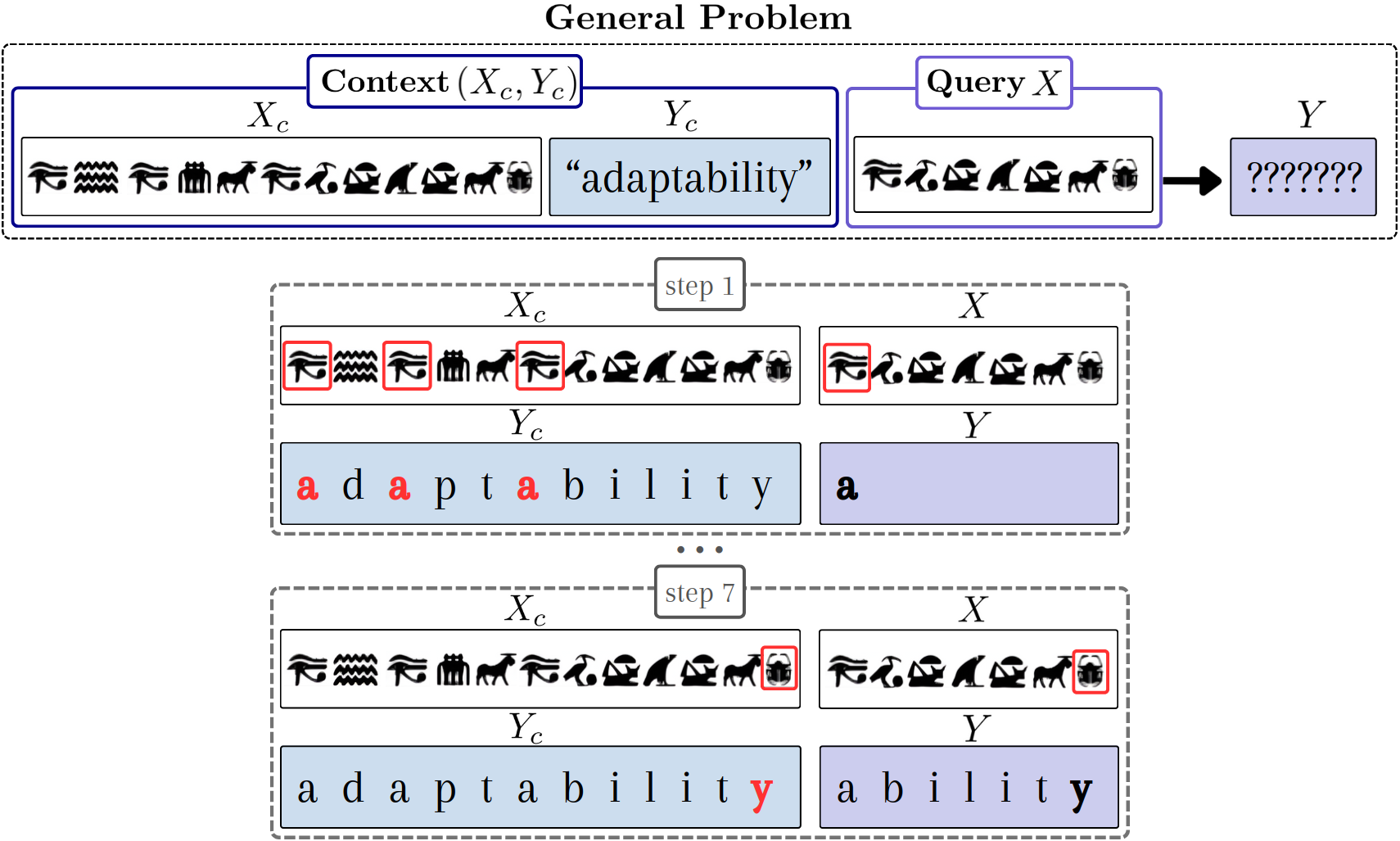}}
  \caption{To classify sequences of unknown symbols in a query image \(X\), Rosetta leverages a context image \(X_c\) containing similar symbols along with their associated labels in \(Y_c\). At each step during decoding, Rosetta identifies matching symbols in the context image and assigns the corresponding label provided in the textual context, highlighted in red in the figure.}
  \label{fig:in-context}
\end{figure}

We introduce \textbf{Rosetta}, a multimodal model designed to leverage contextual information for classifying sequences of novel text and symbol patterns. As illustrated in Figure~\ref{fig:in-context}, by dynamically adapting to a provided context, Rosetta can classify previously unseen patterns, even under significant distribution shifts, thereby eliminating the need for explicit retraining. Rosetta is explicitly trained to utilize context without requiring a language model.

\vspace{-5mm}

Rather than learning static associations between visual patterns and labels, Rosetta is trained with a \textbf{Context-Driven Classification Paradigm}, where the label assigned to a given symbol sequence is defined by its encoding within the context. This enables the model to assign different labels to the same visual pattern depending on the provided context.
This approach enables the classification of novel text and symbol patterns while eliminating the need for explicit retraining.

We enable the application of this training paradigm by introducing a \textbf{Context-Aware Tokenizer (CAT)}. Unlike conventional tokenizers that assign fixed labels to symbols, CAT dynamically encodes label tokens based on the provided context, specifically by dynamically encoding tokens relative to their positions in the context. Another significant advantage of the Context-Aware Tokenizer is its Open-Vocabulary Classification ability, i.e, the prediction of classes that were not encountered during training. By converting any character string into a sequence of unique tokens and then reconstructing the character string based on the original character composition, CAT enables classification across an unlimited range of classes. This capability enables generalization to new alphabets and languages, making it a highly versatile tool. 

Training Rosetta to exploit the image context only requires a specific dataset free of language biases. To this end, we have developed a \textbf{controlled dataset generation strategy} with three key objectives. First,  to ensure that the model relies exclusively on context-driven predictions, rather than drawing on linguistic knowledge, we generate query images composed of random character sequences. Second, we design the dataset to allow in-context learning by maintaining a strong visual link between query and context images, allowing the model to effectively utilize contextual information. Finally, we expose the model to a diverse range of contextual scenarios, varying the informativeness of the context to better simulate real-world recognition challenges. To summarize, our key contributions are as follows:
\begin{enumerate}
    \item \textbf{Introduction of a Context-Driven Classification Paradigm}: To address the limitations of classifier models, such as OCR systems, in recognizing unknown text and symbol patterns, we introduce a context-driven classification paradigm.  Instead of memorizing fixed associations between visual patterns and labels, the model learns to classify a pattern based on how it is encoded within a given context. As a result, it can generalize to unseen text and symbols by performing context-driven predictions.

    \item \textbf{Introduction of the Rosetta Architecture and the Context-Aware Tokenizer (CAT):} We introduce Rosetta, a multimodal architecture designed to classify arbitrary text and symbol patterns by leveraging contextual information. At the core of Rosetta lies the \textbf{Context-Aware Tokenizer (CAT)}, which enables the application of the Context-Driven Classification Paradigm. This design unlocks \textbf{Open-Vocabulary Classification}, allowing the model to classify previously unseen classes and generalize to new alphabets and languages.

    \item \textbf{Exploration of In-Context Learning (ICL) without LLM :} We investigate the feasibility of performing ICL without relying on a large language model or explicit linguistic understanding. This approach enables the development of a model capable of classifying novel text and symbol patterns without requiring retraining, thereby enhancing flexibility and generalization in recognition tasks.
    
    \item \textbf{Development of a dataset generation strategy designed for contextual learning:} We design a data generation strategy that ensures context-driven predictions by applying a linguistic neutrality, establishes a strong visual link between context and query images, and incorporates diverse contextual scenarios. 
\end{enumerate}

\section{Related Work}

 \subsection{Limitations of OCR Models for Unseen Text and Symbols}

Optical Character Recognition (OCR) is a well-established field focused on converting textual images into machine-readable formats. The field has undergone a major transformation with the advent of Transformers \cite{transformers}, leading to Transformer-based architectures such as \cite{dessurt, dan, trocr}. More recently, the rise of Large Language Models (LLMs), including \cite{gpt3, flan, opt, llama, llama2} has further revolutionized OCR systems.  
LLM-enhanced OCR architectures such as \cite{qwen2-vl, got, donut, ureader, daniel, blip2, llava} have significantly expanded OCR applications beyond traditional text recognition. 
These models have improved the ability to process various types of images, including natural scenes, charts, tables, web pages, and handwritten documents, improving their adaptability to various document formats. Moreover, they have enabled true multitasking by handling a wide range of complex text-related tasks, such as Scene Text Recognition, Fine-Grained OCR, Visual Question Answering (VQA) and more. This shift marks a transition toward more flexible, multimodal OCR systems capable of addressing diverse real-world document processing challenges.  

Despite these advancements, OCR models continue to face inherent limitations. Their increasing model size enhances adaptability but introduces extensive computational and data requirements. Furthermore, their robustness remains constrained by reliance on predefined languages, sensitivity to font and style variations. Consequently, these models struggle with out-of-distribution (OOD) data, such as unseen text and symbol patterns or linguistic shifts, necessitating frequent fine-tuning and dataset augmentation to maintain generalization and robustness. Since these models are trained to memorize fixed associations between visual patterns and class labels, their adaptability to new patterns with significant distribution shifts is limited.
Another key limitation is that these models are inherently restricted to predicting only the classes they were trained on, which becomes particularly critical when handling unfamiliar alphabets or languages. To address these limitations, we investigated a recent training paradigm: the~\textbf{In-Context Learning} (ICL).

\subsection{Multimodal In-Context Learning (MICL)}

Recently, leveraging context to adapt to new tasks or enhance performance without fine-tuning has become a key research topic, commonly referred to as In-Context Learning (ICL). Specifically, the model receives a prompt containing a series of inputs with their respective ground truth (i.e. few shot) that illustrate a given task. By leveraging this contextual information, the model can generalize and generate accurate outputs without requiring additional training, thereby offering a flexible and efficient alternative to traditional fine-tuning methods. This adaptation capability is particularly promising for vision-related tasks, including Pattern Recognition (PR) and Optical Character Recognition (OCR).  

ICL was initially introduced in Large Language Models (LLMs) \cite{gpt3,icl_explanation} for textual data. Building on this success, multimodal models have been developed to integrate in-context examples using both images and texts, thereby introducing the Multimodal-ICL (MICL).  Multimodal Architectures such as \cite{flamingo,openflamingo,otter,idefics,idefics2,kosmos,mmicl,emu2,mm1,blip3} allow MICL, while others like \cite{donut,ureader,daniel,blip2,llava} are limited to processing a single image at a time. Multimodal context is mainly incorporated through two architectural approaches: masked cross-attention \cite{flamingo,openflamingo,otter,idefics,idefics2} and decoder-only \cite{kosmos,mmicl,emu2,mm1,blip3} designs. However, further research is needed to determine the most effective approach for context integration.

Studies on Multimodal In-Context Learning (MICL) models, such as \cite{flamingo,idefics}, highlight several limitations in fully utilizing multimodal information. In \cite{whatmakes}, the authors demonstrate that existing models tend to rely predominantly on textual inputs, leading to suboptimal exploitation of visual information. Similarly, \cite{introspection} emphasizes that the success of Multimodal In-Context Learning depends on the meaningful and complementary integration of both modalities. Additionally, \cite{link} underscores the importance of establishing strong connections between the context and query data, introducing the concept of Link-Context Learning.

Despite these challenges, recent models such as \cite{idefics2,mm1,blip3} have improved adaptability by leveraging multimodal context more effectively. This is particularly relevant for document understanding tasks, such as VQA on datasets like OCR-VQA \cite{ocrvqa}, OKVQA \cite{okvqa}, DocVQA \cite{docvqa}. These improvements stem primarily from advancements in training methodologies, particularly in dataset construction. In \cite{otter}, the authors demonstrated the utility of instruction-based formats, establishing stronger connections between images and text pairs—examples. Recent research \cite{idefics2,mm1} demonstrates that multimodal interleaved datasets, i.e., the alternating sequence of images and text, play a crucial role in enabling MICL. Additionally, synthetic data has been shown to enhance in-context few-shot capabilities \cite{e2str,mm1}.

As larger architectures exhibit stronger in-context few-shot capabilities \cite{gpt3,icl_explanation,flamingo}, it has encouraged the development of very large models.  To date, little research has explored cost-efficient approaches to multimodal in-context learning. While MICL has primarily been applied to large-scale models, the idea of leveraging multimodal context to enhance the adaptability of smaller models remains under-explored. To the best of our knowledge, only E2STR \cite{e2str} has attempted MICL for scene text recognition (STR) reproducing Flamingo architecture but using a small-scale language model with only 125M parameters \cite{opt}. This work suggests that there may be viable strategies for enabling MICL adaptation in lightweight models, motivating further investigation into cost-efficient MICL techniques.

Aware of the limitations of OCR models in processing unknown languages, text, and symbol patterns, and motivated by the growing impact of In-Context Learning (ICL), we introduce \textbf{Rosetta} : an alternative approach for open-vocabulary text and symbol pattern classification.

\section{Classifying the Unknown with Rosetta}

We propose a context-driven classification paradigm that enables the classification of unknown text and symbol patterns. This approach mirrors how humans, historically, have interpreted unknown symbols—similar to how the Rosetta Stone was used to decode Egyptian hieroglyphs—by drawing analogies with familiar references \cite{rosetta}. To implement this training paradigm, we present Rosetta, a multimodal architecture. A central component of the Rosetta architecture is the Context-Aware Tokenizer (CAT), detailed in Section 3.1,  which enables this context-driven classification approach. Its design unlocks Open-Vocabulary Classification, allowing the model to classify previously unseen classes and thus adapt to new alphabets and languages. 

\subsection{Context-Driven Classification }

Rather than memorizing fixed associations between symbols and labels, the model is trained to perform context-driven predictions.

\vspace{-5mm}
\begin{figure}[h]
  \centering
  \resizebox{9cm}{!}{\includegraphics{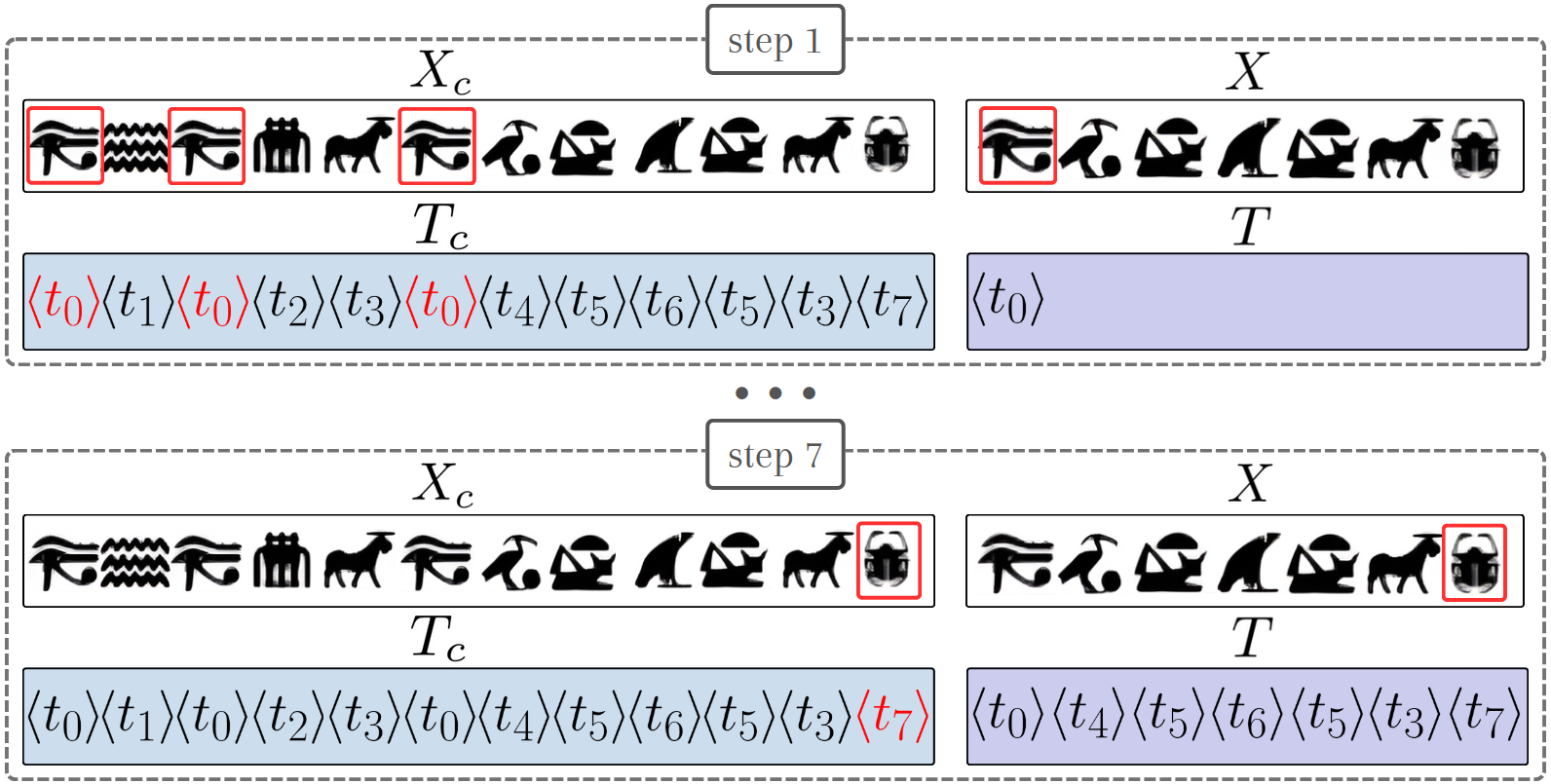}}
  \caption{To classify sequences of unknown symbols in a query image \(X\), Rosetta leverages a context image \(X_c\) containing similar symbols along with their associated labels in \( T_c \). \( T_c \) represents the tokenized encoding of the symbols in \(X_c\), preserving their order of appearance.
  At each step, Rosetta identifies matching symbols in the context image and assigns the corresponding label provided in the textual context, highlighted in red in the figure.}
  \label{fig:context-dependent}
\end{figure}

As illustrated in Fig.~\ref{fig:context-dependent}, given a query image \( X \) containing a sequence of symbols, the model is trained to predict a corresponding sequence of class tokens \( T \) by leveraging a multimodal context \( C = (X_c, T_c) \).  

During training, the ground truth \(T\) associated with the query image \(X\) is generated using the Context-Aware Tokenizer, based on the order of symbol appearance in the context \(C\). Any change in this order leads to a corresponding modification of the ground truth \(T\). Moreover, if a symbol \( s_k \) in \( X \) is absent from \(C\), the ground truth for that symbol in \( T \) is assigned the special token `\(\langle ooc\rangle\)` (out-of-context), indicating that \( s_k \) does not exist in the reference context \(C\).  

Since the model is trained to perform context-driven predictions rather than memorizing fixed symbol-label associations, it can generalize to unseen text and symbols by leveraging contextual information.  

\subsection{Context-Aware Tokenizer (CAT)}
\vspace{-6.5mm}
\begin{figure}[h]
  \centering
  \resizebox{10cm}{!}{\includegraphics{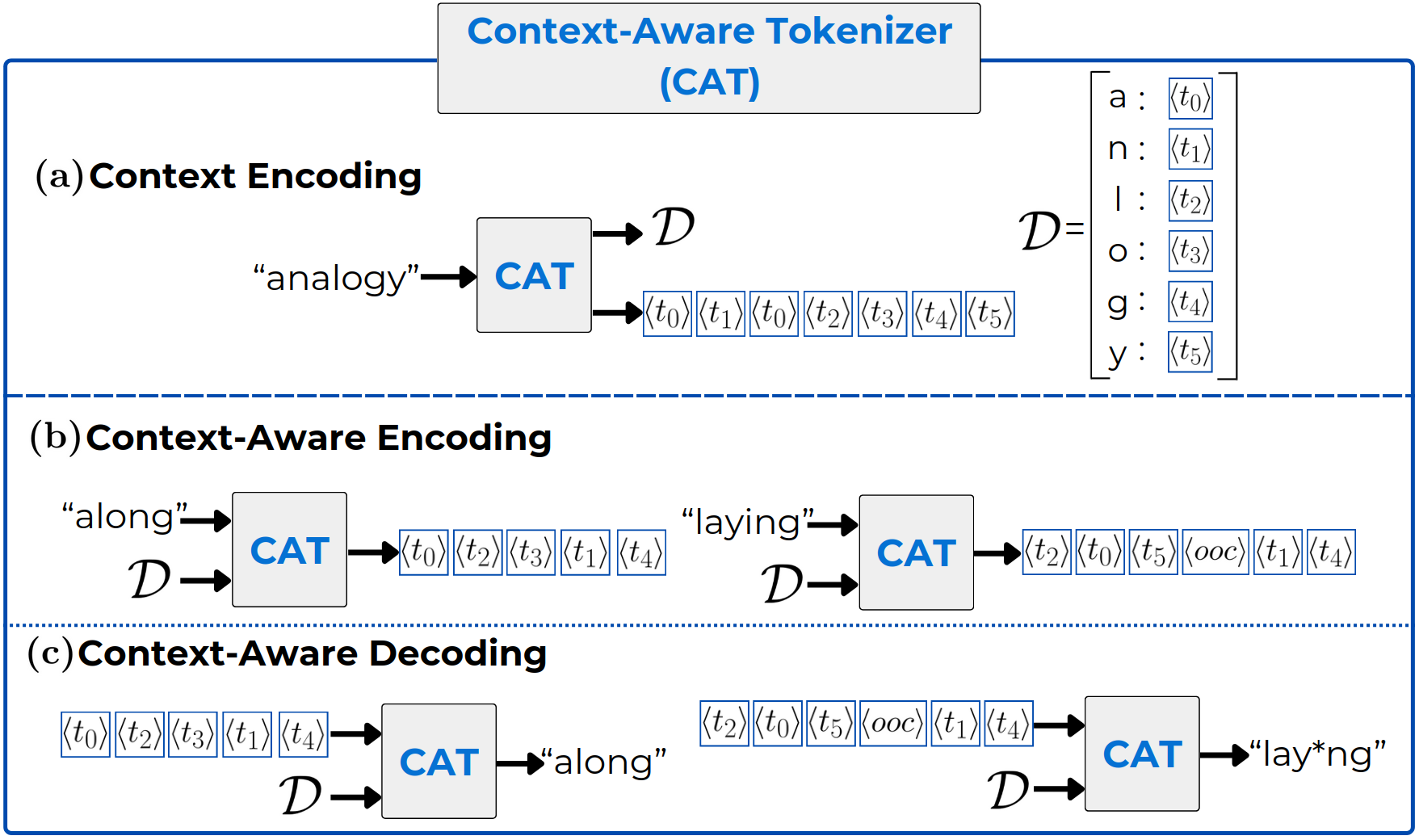}}
\caption{Illustration of context and query text encoding/decoding using the Context-Aware Tokenizer (CAT). (a) CAT encodes a context text and stores all character-token mappings in a dictionary \( \mathcal{D} \). This dictionary is then used by CAT to  (b) encode new text or to (c) decode predictions from the model. The '*' character denotes predictions corresponding to the \( \langle ooc \rangle \) (out-of-context) token.}
  \label{fig:cat}
\end{figure}
\vspace{-2mm}

To achieve context-driven classification of text and symbol patterns, we introduce a specialized tokenizer: the Context-Aware Tokenizer (CAT). As illustrated in Fig. \ref{fig:cat}, unlike traditional static tokenization methods, CAT constructs a dynamic context-dependent mapping that ensures consistent tokenization of the query text in accordance with the context. 

\subsubsection{(a) Context Encoding}
The Context-Aware Tokenizer (CAT) encodes the sequence of characters \(Y_c\) into a sequence of tokens \( T_c = \{ \langle t_0 \rangle,  \dots, \langle t_L \rangle \} \) by assigning tokens based on their positions and their order of appearance. The first unique character is assigned the token \( \langle t_0 \rangle \). Each new unique character gets a new token \( \langle t_k \rangle \). If a character has already appeared in the sequence \(Y_c\), it is assigned the same token as its first occurrence. At the end of this process, all character-token pairs are stored in a dictionary \( \mathcal{D} \).

\subsubsection{(b) Context-Aware Encoding}
To encode a sequence of characters \(Y\) into a sequence of tokens \( T \), which is used as the ground truth during training, CAT leverages the dictionary \( \mathcal{D} \) of character-token associations built during the context encoding. If a character does not appear in \( \mathcal{D} \), the token \( \langle ooc \rangle \) (out-of-context) is assigned.

\subsubsection{(c) Context-Aware Decoding}
To decode a sequence of tokens \( T \) back into a sequence of characters, CAT uses the dictionary \( \mathcal{D} \) built during context encoding. In this case, the token \( \langle ooc \rangle \) is replaced by a special character, which we select as '*'.

A key advantage of the Context-Aware Tokenizer is that it enables the prediction of classes that were not encountered during training. Since the CAT converts any character string into a series of unique tokens and then decodes the model’s predictions back into a character string based on the contextual text, it allows the classification of text and symbol patterns across an unlimited range of classes.  The vocabulary size of CAT can be adjusted based on the maximum context size to be processed, specifically according to the maximum number of unique symbols within the context.

\subsection{Rosetta Architecture}
\begin{figure}[h]
  \centering
  \resizebox{\textwidth}{!}{\includegraphics{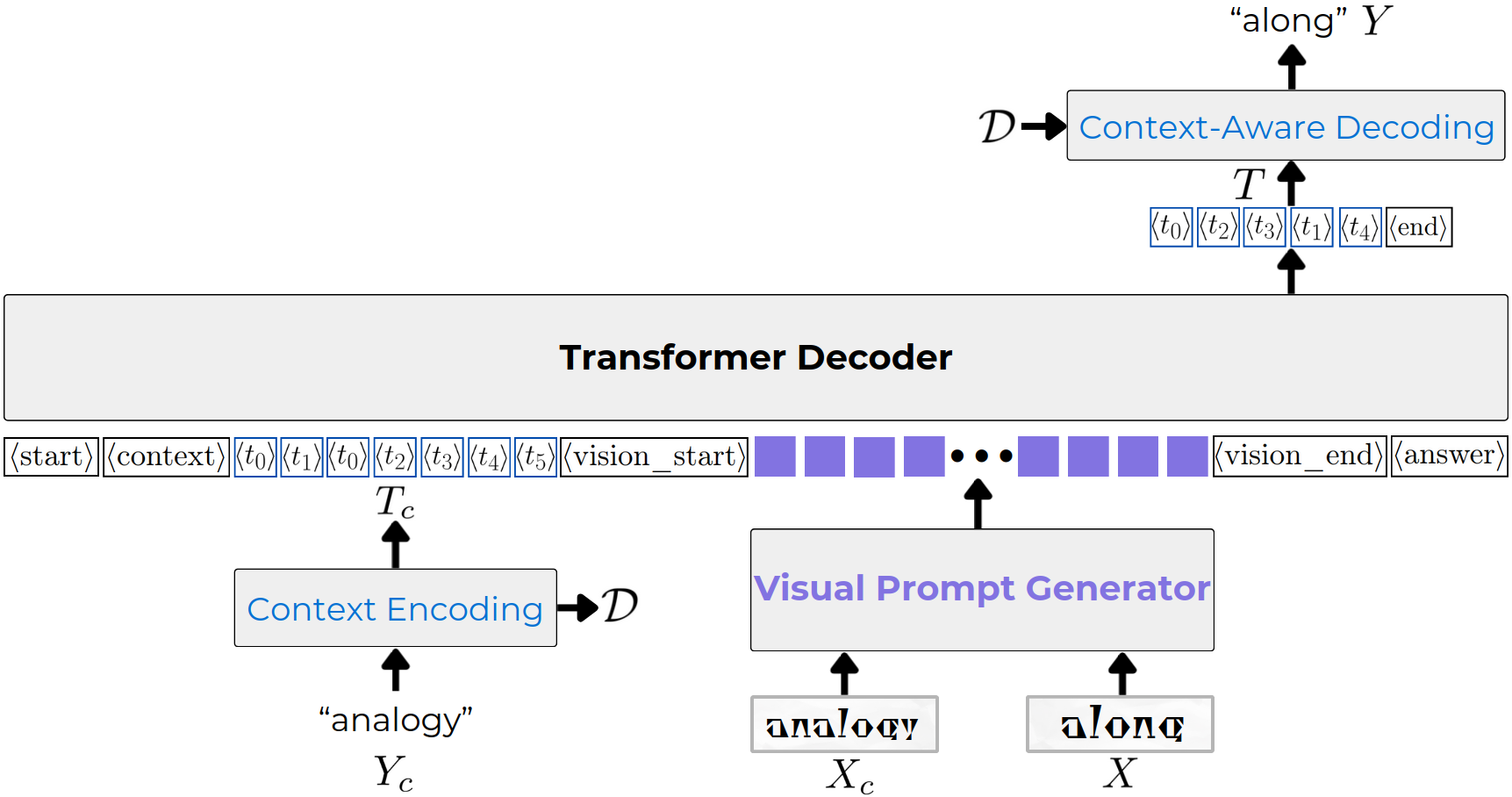}}
\caption{Illustration of the Rosetta architecture, structured around three core components: (1) a Context-Aware Tokenizer (CAT) that encodes the context text \(Y_c\) into a sequence of tokens \(T_c\) and decodes the predicted sequence of tokens \(T\) back into a sequence of characters \(Y\) using a dictionary \(D\) of character-token association; (2) a Visual Prompt Generator (VPG) that converts the context and query images (\(X_c\), \(X\)) into token sequences interpretable by the transformer decoder; and (3) a transformer decoder that processes the multimodal data from both the CAT and the VPG to predict \(T\), a sequence of tokens corresponding to the symbols in the query image.}

  \label{fig:architecture}
\end{figure}

The Rosetta architecture, illustrated in Fig.\ref{fig:architecture},  is designed for processing text and images, and is structured around three core components: (1) a context-aware tokenizer, (2) a visual prompt generator, and (3) a transformer decoder.

The \textbf{Context-Aware Tokenizer (CAT)}, detailed in Section 3.2, dynamically encodes the context text \(Y_c\) to a sequence of tokens \(T_c\) and finally decodes the predicted sequence of tokens \(T\) to a sequence of characters \(Y\). 

The \textbf{Visual Prompt Generator (VPG)} follows a similar approach to the image encoding mechanism used in Qwen2-VL \cite{qwen2-vl}. VPG converts the context and query images (\(X_c\), \(X\) ), regardless of their resolution, into token sequences that are interpretable by the transformer decoder. Initially, the images are stacked and transformed into variable-length visual tokens. In contrast to the approach in Qwen2-VL, which utilizes a 3D convolution between two video frames, we apply this method to the context and query images. This modification reduces the number of patches while simultaneously establishing an initial link between the context and query images within the architecture. The patches are then processed using a Vision Transformer \cite{vit} with a 2D Rotary Positional Embedding (2D-RoPE) \cite{rope}, which is crucial for capturing the two-dimensional positional information of the images. Lastly, the generated tokens are projected through linear layers to match the embedding size required by the transformer decoder.

After being processed by the CAT and VPG, text and image tokens are separated with special tokens ( \(\langle \text{vision\_start} \rangle,  \langle \text{vision\_end} \rangle\)) and a Multimodal RoPE (MRoPE) \cite{qwen2-vl} is applied before feeding them into the transformer decoder. The \textbf{Transformer Decoder} consists of causal self-attention layers, feed-forward MLPs, and layer normalization components. It processes the multimodal data from both the CAT and the VPG to predict a sequence of tokens corresponding to the symbols in the query image.
Finally, CAT decodes the token sequence \(T\) and maps it back to \(Y\) using the original alphabet employed in \(Y_c\).

\section{Experimental Setup and Results}

\subsection{Training Dataset}
During training, synthetic data is dynamically generated, featuring context-query image pairs (\( X_c, X \)) and their corresponding transcriptions (\( Y_c, Y \)). The textual transcriptions (\( Y_c, Y \)) are then encoded by the CAT into token sequences (\( T_c, T \)). Recall that the training objective is to predict \(T\) based on \(X\) and a context \(C = (X_c, T_c)\) (see Fig. \ref{fig:context-dependent}).
To ensure that Rosetta relies exclusively on context-driven predictions, rather than drawing on linguistic knowledge, we randomly sample a sequence  \(Y\) ranging from 1 to 15 characters, maintaining linguistic neutrality.
Followed by random selection of a font \( f \) from the font set \(  \mathcal{F} \). The query image \( X \) is then rendered using this query text \(Y\) and the selected font~\( f \). The use of the same font  \( f \) establishes a strong visual link between the query and context images, enabling the model to effectively leverage contextual information.

\begin{figure}[h]
  \centering
  \resizebox{9.5cm}{!}{\includegraphics{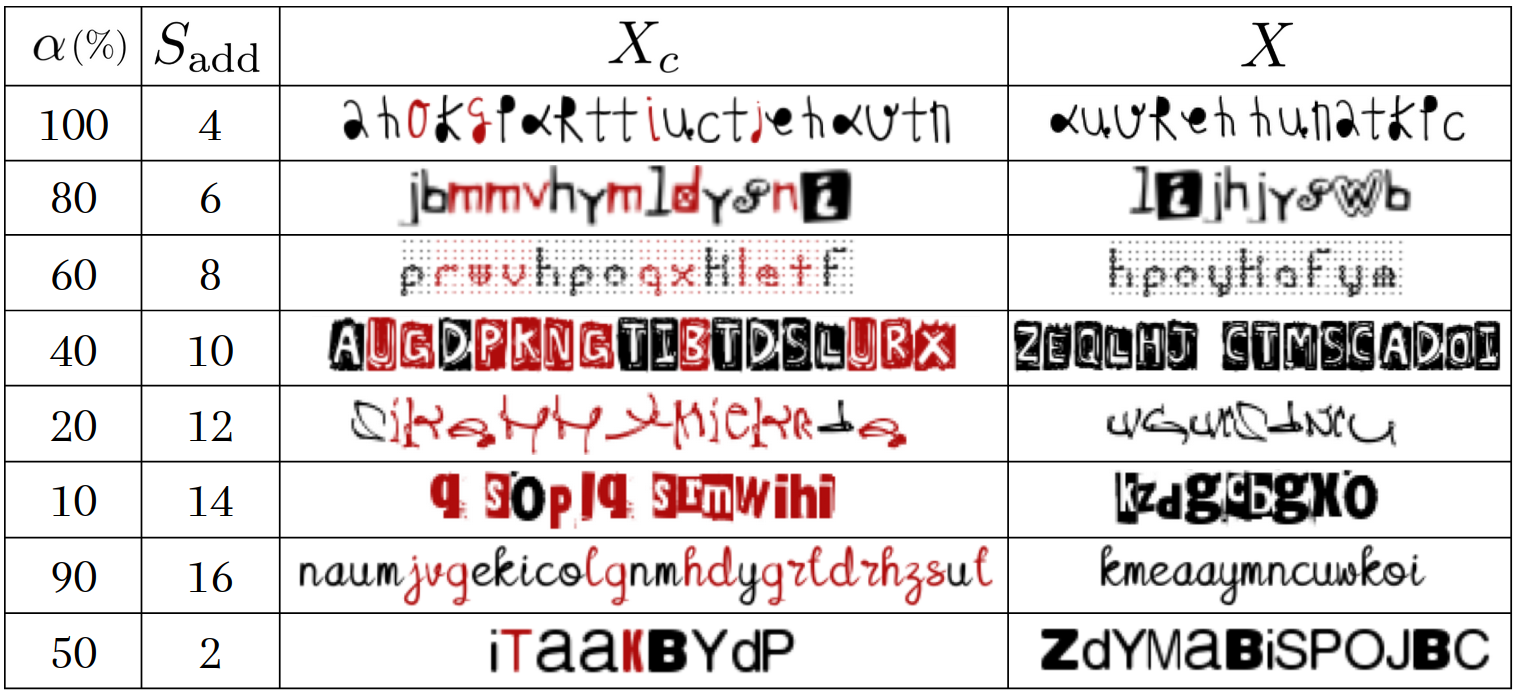}}
  \caption{Samples of context and query images from the training set, showing variations in the coverage rate \(\alpha\) and the number of symbols \( S_{\text{add}} \). The red symbols in \(X_c\) represent symbols that belong to \( S_{\text{add}} \). The red color is used for illustration purposes only.}

  \label{fig:train_data}
\end{figure}

We employed a collection of 4,000 distinct open-source fonts\footnote{The fonts used for this paper are open-source and downloadable from \href{https://www.dafont.com}{dafont.com}} covering a diverse range of cursive, non-cursive styles and decorative text styles (see Fig.\ref{fig:train_data}). This selection ensures exposure to various typographic patterns, enhancing the model's ability to generalize across different text appearances. For the experiments in this paper, the set of characters \(\mathcal{C}\) used during training is restricted to lowercase Latin letters. 

To ensure that the model is trained under diverse conditions, where the contextual information available for decoding \(X\) ranges from highly informative to ambiguous, a context text \(X_c\) is generated using 2 parameters that control the data variability:
\begin{itemize}
    \item \textbf{Query coverage rate \(\alpha\)}  : Randomly selected between 0\% and 100\% of the set of unique symbols required to decode \(X\).
    \item \textbf{Added symbols  \( S_{\text{add}}\) }: Ranges from 0 to 20 additional irrelevant symbols from  \( \mathcal{C} \setminus X \), which are not necessary to decode \(X\).
\end{itemize}

Samples of generated data following that process can be found Fig. \ref{fig:train_data}. 

\vspace{-2mm}
\vspace{-5mm}

\subsection{Evaluation Datasets}  
\begin{figure}[h]
  \centering
  \resizebox{\textwidth}{!}{\includegraphics{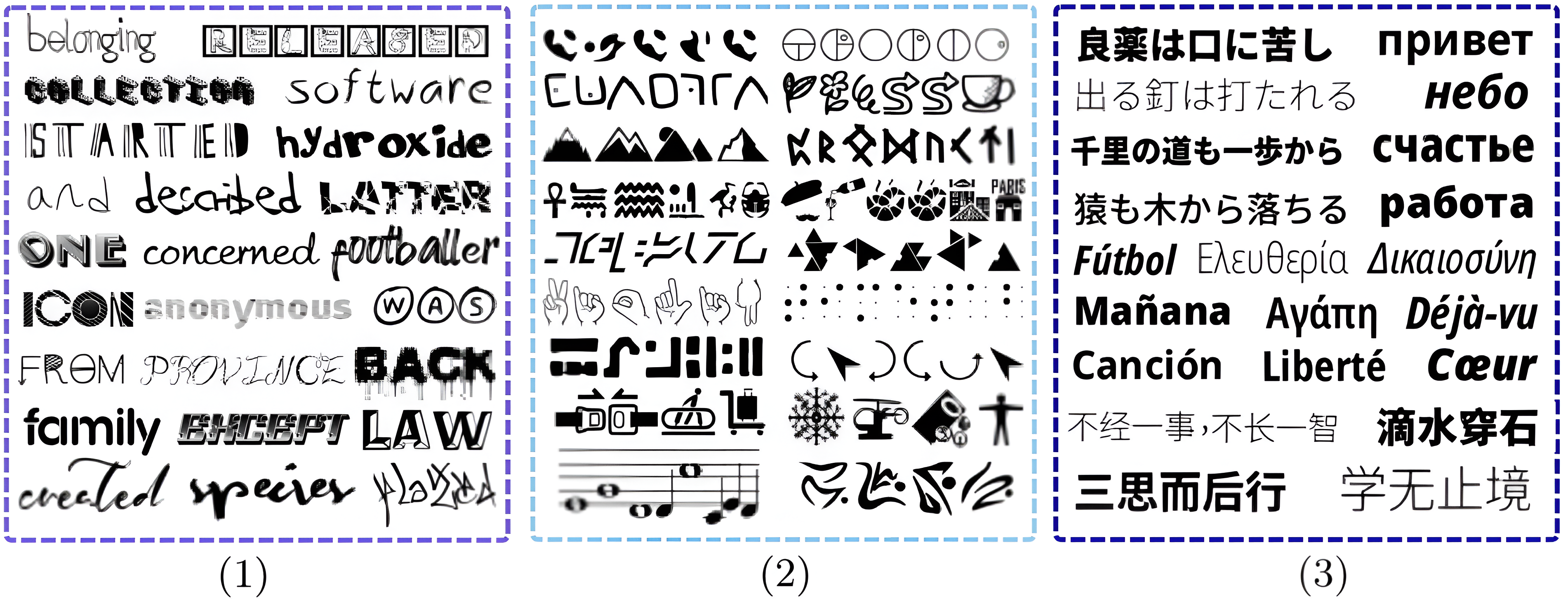}}
  \caption{Samples of query images \( X \) from the test sets: (1) unseen textual patterns, (2) unseen symbols (3) multi-alphabet.}
  \label{fig:test_sets}
\end{figure}

To systematically evaluate the generalization capabilities of our model, we designed three distinct test sets, each targeting a specific challenge: unseen text patterns, out-of-distribution (OOD) symbol patterns, and adaptation to entirely new alphabets. These datasets allow us to assess the model's robustness across different typographic variations, symbol structures, and linguistic systems. Samples from the evaluation datasets are illustrated on  Fig. \ref{fig:test_sets}. The evaluation sets are structured as follows:

\begin{itemize}
    \item \textbf{Test 1: Unseen Text Patterns, Latin Alphabet} (See Fig. ~\ref{fig:test_sets}.1):  
    This test set evaluates the model’s ability to generalize to novel text patterns. This dataset includes 470 unseen fonts that preserve traditional character structures while introducing new typographic variations. The text samples are composed of real English words extracted from Wikipedia. Recall that during training no English word was seen as character sequences are randomly generated. Real words are introduced here only to facilitate the qualitative review of the tests results.

    \item \textbf{Test 2: Unseen Symbol Patterns, Latin Alphabet} (See Fig. ~\ref{fig:test_sets}.2): 
    This test set is designed to evaluate the model's ability to classify previously unseen symbols. The dataset employs 100 fonts that transform each Latin alphabet character into an abstract symbolic pattern, eliminating any recognizable character structure. As a result, the symbols are entirely out-of-distribution (OOD) relative to standard character recognition. The text samples are also composed of real English words extracted from Wikipedia. 

    \item \textbf{Test 3: Unseen Text and Symbol Patterns, New Alphabets} (See Fig.~\ref{fig:test_sets}.3): 
    This test assesses the model’s ability to generalize beyond its original training distribution by handling both unseen character classes and new writing systems.  This dataset includes proverbs in Japanese and Chinese, as well as words in Greek, Russian, Spanish, and French. Each language is represented in its native script (e.g., Greek words in Greek characters), ensuring an authentic evaluation of cross-lingual adaptability. The dataset employs 9 fonts per language.
\end{itemize}

\vspace{-8mm} % Reduce space 

All the test sets are generated offline to ensure consistency across evaluation runs. All the queries to decode contain between 1 and 15 symbols. The context is generated following the same process as during training, i.e., with a query coverage rate between \(0\%\) and \(100\%\), and an additional number of irrelevant symbols \( S_{\text{add}} \) ranging from 0 to 20.

\subsection{Comparison Between an OCR-Based Model and Rosetta}
To evaluate Rosetta's context-driven classification paradigm in comparison to a conventional OCR training approach, we trained a model that closely follows the Rosetta architecture. The goal is to compare the training paradigms by evaluating two models with a similar number of parameters, both exposed to the same range of visual patterns during training. However, instead of utilizing a Context-Aware Tokenizer (CAT), the OCR-based model employs a conventional static tokenizer with a predefined vocabulary consisting of the 26 Latin alphabet characters, without any contextual information. The OCR-based model was trained on the same training set of 4000 fonts to ensure that it encountered a comparable range of typographic variations.  Since the character sequences are randomly organized, neither model can leverage linguistic knowledge to enhance their predictions.

Implementation details concerning the model parameters and data can be found in the appendix.

\subsection{Results}  
To assess the model's ability to leverage multimodal context under varying scenarios, we evaluate its performance on the three test sets described in Section~4.2. For this evaluation, we employ two primary metrics:

First, to evaluate the model in a conventional text classification setting —predicting the character class of each symbol—we compute the Character Error Rate (CER). In this case, the prediction of an \( \langle ooc \rangle \) token is considered an error.

Second, to measure the model's ability to reproduce the training task — predicting a sequence of tokens where each token corresponds to the label of a symbol if it is present in the context, or predicting an \( \langle ooc \rangle \) token otherwise— we compute the Token Error Rate (TER). The TER is defined as the edit distance between the predicted and ground-truth token sequences, normalized by the total number of tokens.  This metric assesses the increased difficulty in recognizing that the information required to decode a symbol is absent from the context. It is similar to the reject option in a classic classification system.

To evaluate the performance of the Rosetta model, we analyzed the evolution of these metrics under varying context configurations. Specifically, we focused on two aspects: the Query Coverage Rate \(\alpha\), where a 100\% coverage indicates that all symbols necessary for decoding the query are present in the context, and the Irrelevant Symbols Rate \(\beta\) in the context. Formally, \(\beta\) quantifies the proportion of context symbols that do not appear in the query, computed as the number of such symbols divided by the total number of symbols in the context.

\subsubsection{Comparison Between the OCR-Based Model and Rosetta}  
\begin{figure}[h]
  \centering
  \resizebox{7cm}{!}{\includegraphics{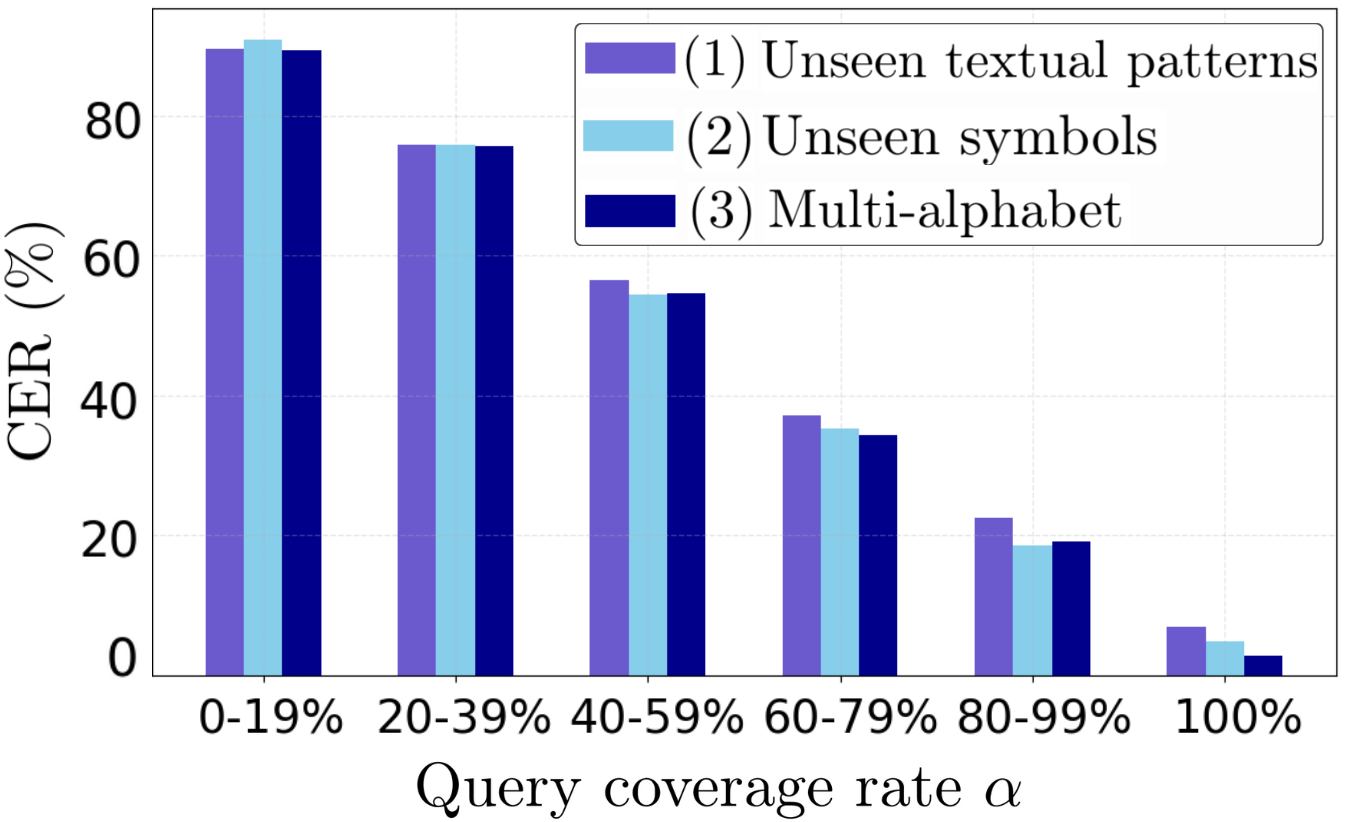}}
  \caption{Evaluation of Rosetta's Character Error Rate (CER) on three test sets under diverse context conditions, varying the query coverage rate, and treating \( \langle ooc \rangle \)  token predictions as errors.}
  \label{fig:cer_alpha}
\end{figure}

 To compare the performance of the OCR-based model with Rosetta on query images from the textual pattern set (see Fig.\ref{fig:test_sets}), we generated a context \(C\) for each query \(X\) with a \(\alpha\) of 100\%, ensuring that both models have to predict the same ground truth. The CER is computed for each individual prediction, and the CER distribution for each model is presented in Table \ref{fig:cer_alpha}. While both models achieve a median CER of 0.0\%—indicating that they generally perform well—Rosetta exhibits a slight performance advantage. Specifically, Rosetta attains a lower mean \textbf{CER of 6.54\%}, compared to 13.56\% for the OCR-based model. Furthermore, the CER distribution is more concentrated around 0.0\% for Rosetta, suggesting greater consistency in classification accuracy. This trend is further reinforced by analyzing the upper percentiles (70\textsuperscript{th} to 100\textsuperscript{th}), which provide insight into how the models handle more challenging cases. A lower CER in these upper percentiles indicates that Rosetta is better at maintaining reliable performance on difficult images, whereas the OCR-based model shows greater variability. This suggests that Rosetta's context-driven approach allows it to generalize more effectively to ambiguous or degraded samples, where traditional OCR methods tend to struggle.
 These findings highlight the effectiveness of Rosetta’s context-driven classification approach, demonstrating performance comparable to traditional OCR systems when provided with sufficient and relevant contextual information.

\vspace{-5mm}
\begin{table}[htbp]
\centering
\begin{tabular}{l S[table-format=2.2] *{6}{S[table-format=3.2]}} % 6 colonnes de percentiles
\toprule
Model & {Mean CER (\%)} & \multicolumn{6}{c}{Percentiles of CER (\%)} \\
\cmidrule(lr){3-8} % Ajustement du cmidrule
& & {50\textsuperscript{th}} & {60\textsuperscript{th}} & {70\textsuperscript{th}} & {80\textsuperscript{th}} & {90\textsuperscript{th}} & {100\textsuperscript{th}} \\
\midrule
OCR-based & 13.56 & 0.00 & 0.00 & 13.33 & 25.00 & 46.67 & 300.00 \\
Rosetta & \textbf{6.54} & 0.00 & 0.00 & \textbf{0.00} & \textbf{7.69} & \textbf{25.00} & \textbf{142.86} \\
\bottomrule
\end{tabular}
\vspace{2mm} % Ajout d'un espace vertical de 2mm
\caption{Evaluation of mean CER and upper percentile CER distribution for the OCR-based model and Rosetta on the textual test set. The context provided to Rosetta ensures 100\% query coverage and incorporates a random irrelevant symbol rate \(\beta\) ranging from 0\% to 100\%.}

\label{tab:deciles_horizontal_mean_cer}
\end{table}
\vspace{-15mm}

\vspace{-10mm}

\subsubsection{Impact of Context Quality}  
As shown in Fig. \ref{fig:cer_alpha}, for all test sets, the CER, computed by considering the prediction of the \(\langle ooc \rangle\) token as an error, decreases linearly as the query coverage rate \(\alpha\) increases.  

\begin{figure}[h]
  \centering
  \resizebox{\textwidth}{!}{\includegraphics{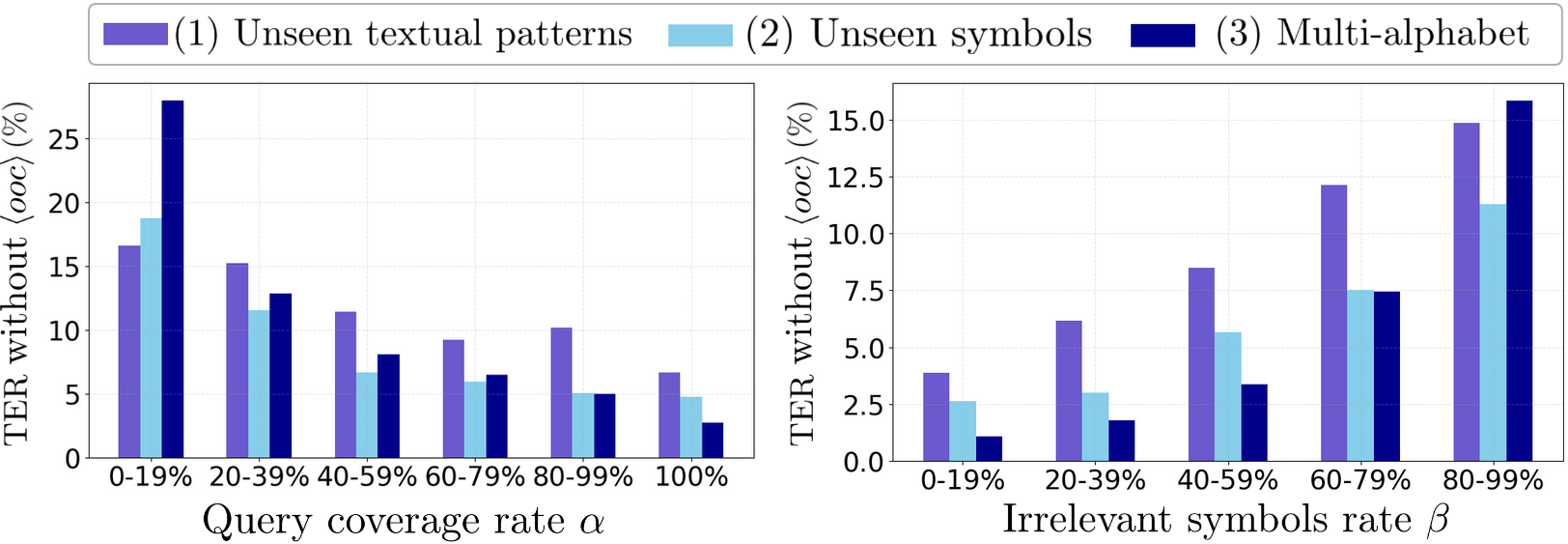}}
  \caption{Token Error Rate (TER) of the Rosetta model, excluding the \( \langle ooc \rangle \) token, across different context scenarios on the three test sets, with variations in \(\alpha\) and \(\beta\).}
  \label{fig:alpha_beta_no_ooc}
\end{figure}

A more detailed analysis of Rosetta's classification performance, excluding the rejection mechanism (\(\langle ooc \rangle\) token prediction) and evaluating the model solely on symbols it can classify based on the provided context, also reveals a strong dependency on context quality.
 As shown in Fig. \ref{fig:alpha_beta_no_ooc}, the TER decreases as the \(\alpha\) increases, indicating that a more informative context enhances classification accuracy. Conversely, a higher rate of irrelevant symbols \(\beta\) in the context leads to an increase in TER, highlighting the negative impact of contextual noise. The more frequently the symbols to be classified appear in the context, the greater Rosetta’s ability to classify them correctly. These findings emphasize the crucial role of high-quality contextual information in maximizing Rosetta’s effectiveness.

\vspace{-5mm}

\vspace{-10mm}

\begin{figure}[h]
  \centering
  \resizebox{\textwidth}{!}{\includegraphics{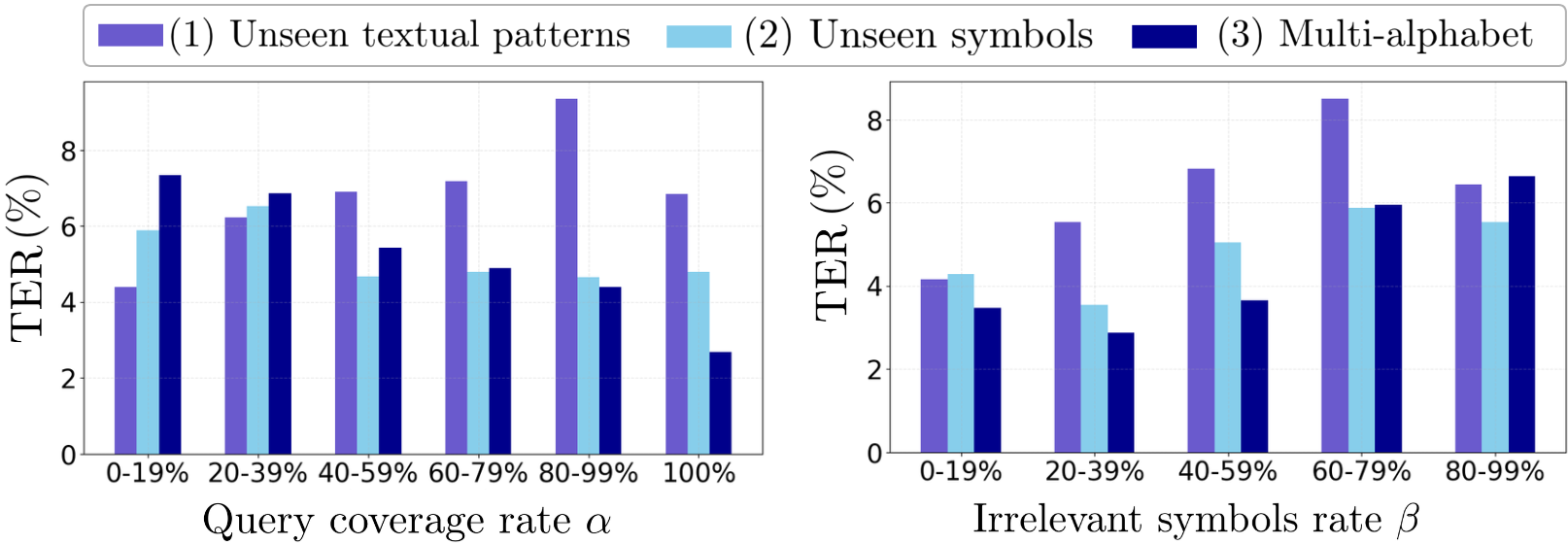}}
  \caption{Token Error Rate (TER) of the Rosetta model across different context scenarios on the three test sets, with variations in \(\alpha\) and \(\beta\).}

  \label{fig:ter_alpha_beta}
\end{figure}

\vspace{-10mm}

\subsubsection{Adaptability To Unknown Text and Symbol and New Alphabets}  
As shown in Figures \ref{fig:cer_alpha}, \ref{fig:alpha_beta_no_ooc}, \ref{fig:ter_alpha_beta}, Rosetta achieves consistent performance across different test sets, regardless of the context scenario. As illustrated Fig. \ref{fig:ter_alpha_beta}, the Token Error Rate (TER), which evaluates the model's ability to correctly classify unknown symbols and determine whether a symbol from the query is present in the context by predicting the \(\langle ooc \rangle\) token, remains consistently below 10\%. These results indicate that Rosetta successfully generalizes to unseen text patterns and symbols beyond those encountered during training. Additionally, the model effectively adapts to new character sets, including previously unseen alphabets, highlighting its potential for multilingual text recognition.

\begin{figure}[h]
  \centering
  \resizebox{\textwidth}{!}{\includegraphics{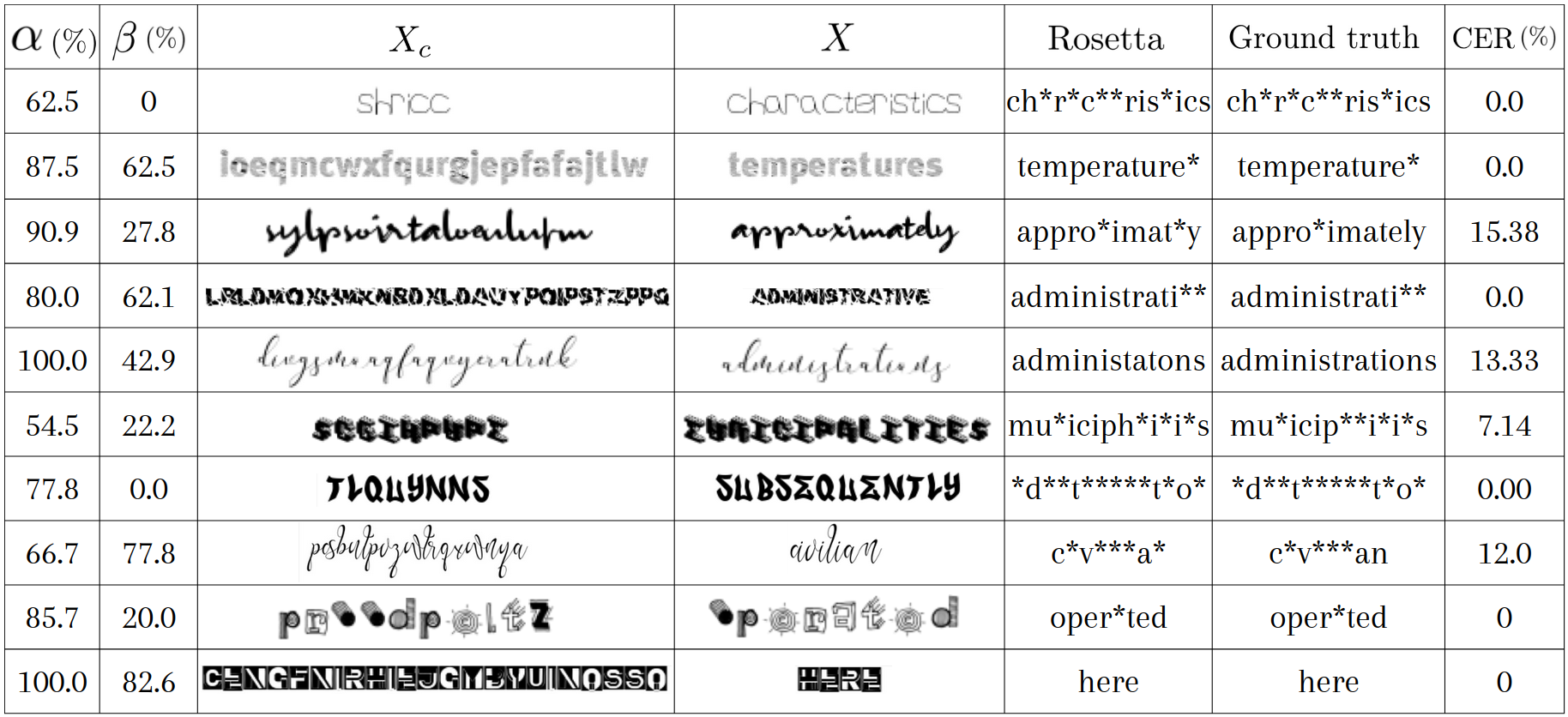}}
  \caption{Qualitative results of Rosetta on the textual test set under different context scenarios with varying query coverage rate  \(\alpha\)  and irrelevant symbol rate \(\beta\).}
  \label{fig:pred_test1}
\end{figure}

\begin{figure}[h]
  \centering
  \resizebox{8.5cm}{!}{\rotatebox{90}{\includegraphics{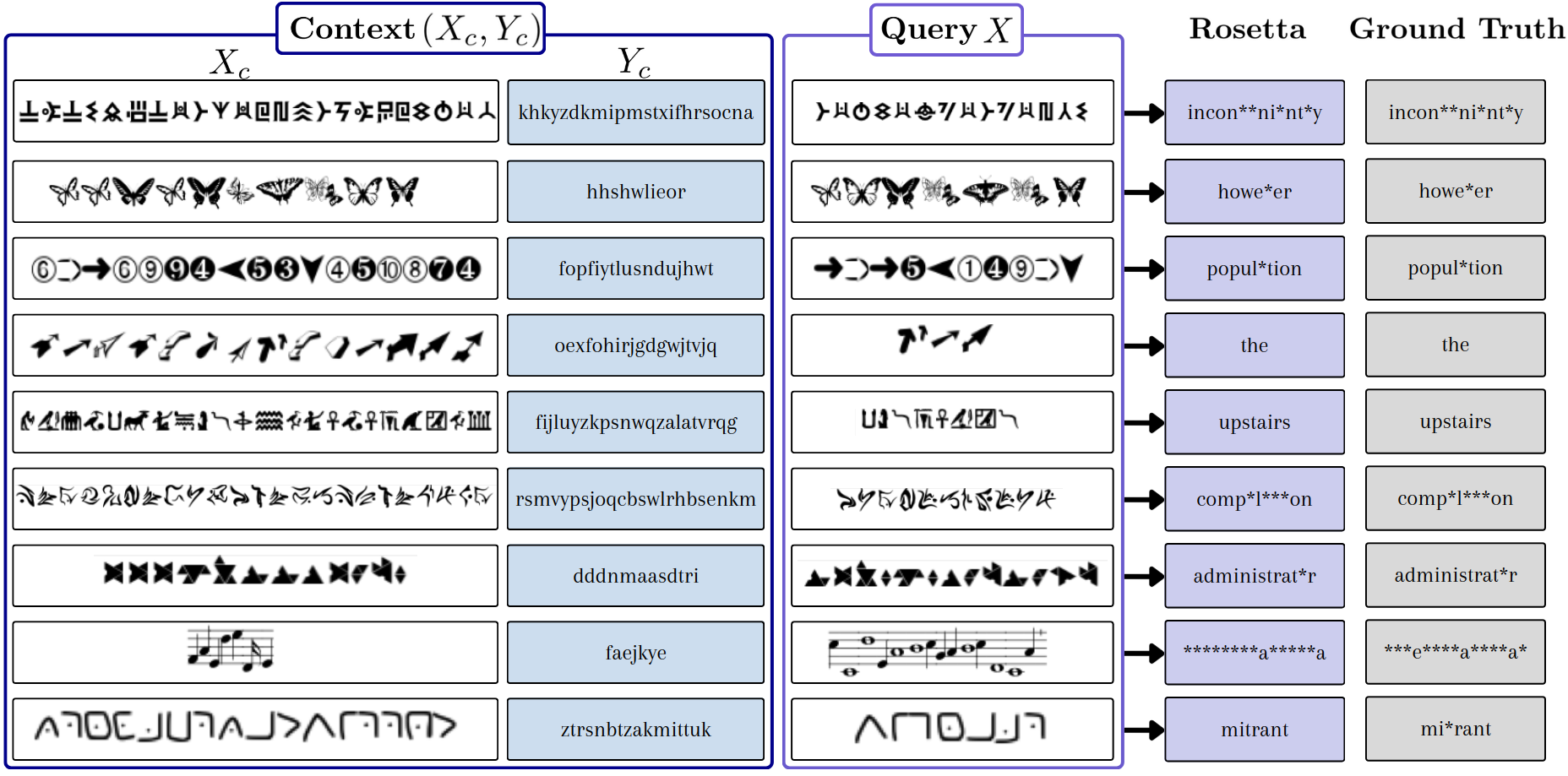}}}
  \caption{Qualitative results of Rosetta on the symbolic test set under different context scenarios.}
  \label{fig:pred_test2}
\end{figure}

\begin{figure}[h]
  \centering
  \resizebox{8.5cm}{!}{\rotatebox{90}{\includegraphics{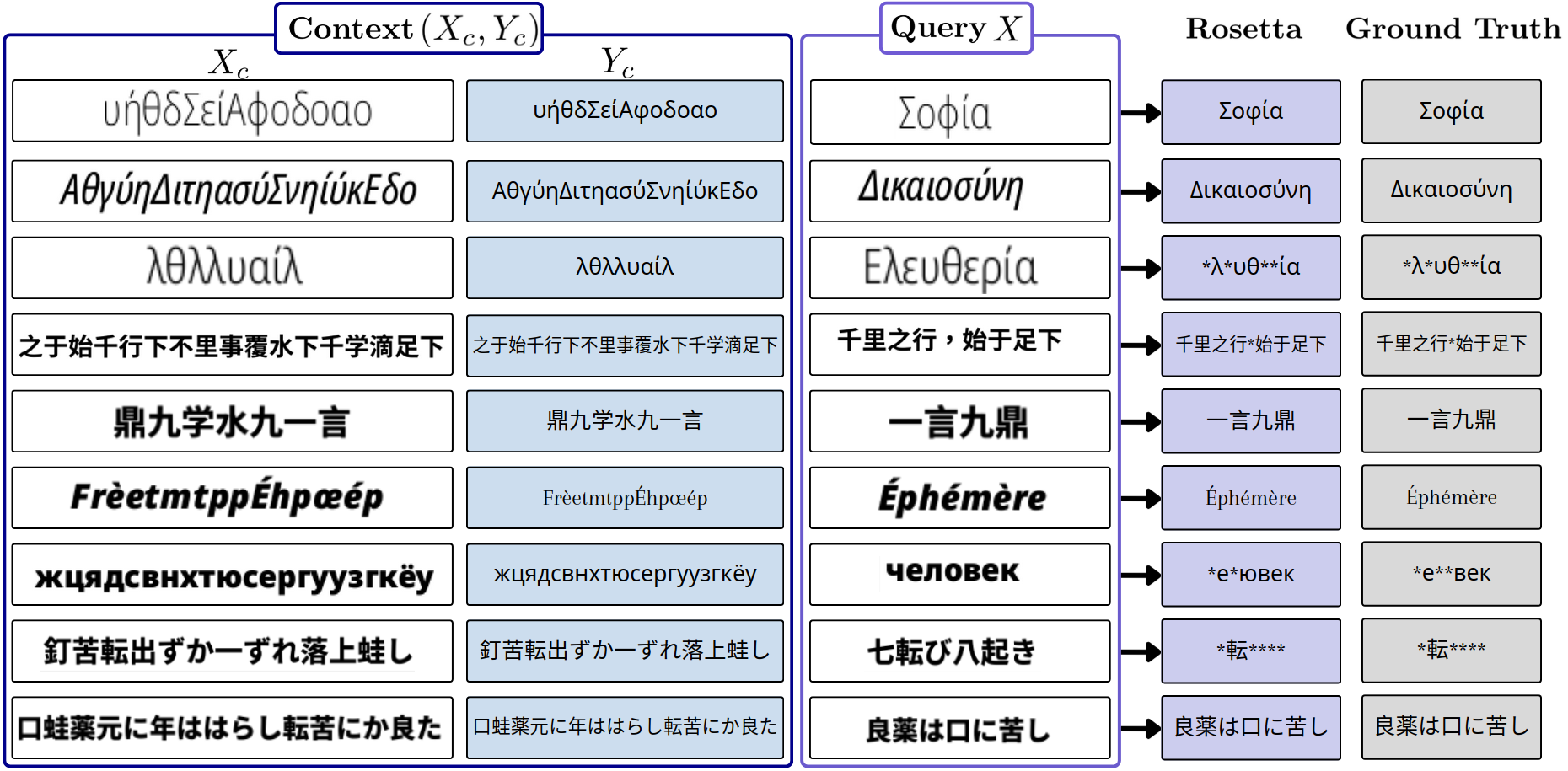}}}
  \caption{Qualitative results of Rosetta on the Multi-Alphabet test set under different context scenarios.}
  \label{fig:pred_test3}
\end{figure}

 Additional quantitative results and qualitative examples of the predictions on the test sets are provided in the appendix, in Figures~\ref{fig:f1_score}, \ref{fig:vs}, \ref{fig:pred_test1}, \ref{fig:pred_test2} and \ref{fig:pred_test3}.

\section{Conclusion}

In this work, we introduced Rosetta, a multimodal model designed for text and symbol classification through in-context learning, without relying on linguistic knowledge. Unlike conventional OCR-based approaches that require extensive retraining when encountering novel text or symbols, Rosetta leverages contextual adaptation to generalize to unseen data.

By proposing a Context-Driven classification paradigm, we enable Rosetta to classify symbols based on their contextual encoding rather than fixed visual-label associations. At the core of this approach lies the Context-Aware Tokenizer (CAT), which dynamically encodes label tokens relative to their context, thereby unlocking Open-Vocabulary classification and, consequently, generalization to new alphabets and languages.

To support effective training, we developed a controlled dataset generation strategy ensuring that Rosetta learns exclusively from contextual cues while being exposed to a diverse range of scenarios to enhance its adaptability. Our results demonstrate that Rosetta successfully classifies unseen text and symbol patterns, even under significant distribution shifts, without the need for retraining.

While our study demonstrates the effectiveness of Rosetta in controlled synthetic settings, several challenges must be addressed for real-world deployment. Future work will focus on optimizing the selection of context examples and expanding Rosetta’s adaptability to a broader range of scenarios, including contexts affected by varying degradation conditions. Additionally, enhancing the model’s scalability and efficiency will be essential to improving its performance on large-scale applications. Beyond strengthening robustness, integrating Rosetta with richer multimodal contexts and exploring its interpretability could further enhance its practical utility. Extending its capabilities to real-world multilingual OCR tasks will be a crucial step toward establishing context-driven recognition as a viable alternative to traditional approaches.

Overall, this study highlights the potential of In-Context Learning (ICL) as a scalable and efficient alternative to supervised learning methods for text and symbol recognition, paving the way for more flexible and adaptive recognition systems.

\subsubsection{\discintname}
The authors have no competing interests to declare that are relevant to the content of this article. 
%\bibliography{mybibliography}

\begin{thebibliography}{8}

\bibitem{qwen2-vl}
Wang, P., Bai, S., Tan, S., et al.: Qwen2-VL: Enhancing vision-language model's perception of the world at any resolution. arXiv preprint \textbf{arXiv:2409.12191} (2024)


\bibitem{transformers}
Vaswani, A.: Attention is all you need. \textit{Advances in Neural Information Processing Systems}, 2017.

\bibitem{rope}
Su, Jianlin and Ahmed, Murtadha and Lu, Yu and Pan, Shengfeng and Bo, Wen and Liu, Yunfeng : Roformer: Enhanced transformer with rotary position embedding. Neurocomputing pages 127063, (2024) Elsevier

\bibitem{vit}
Dosovitskiy, Alexey and Beyer, Lucas and Kolesnikov, Alexander  et al. : An image is worth 16x16 words: Transformers for image recognition at scale (2020) arXiv preprint arXiv:2010.11929


\bibitem{rosetta}
Andrews, C. A. R.: The Rosetta Stone. In \textit{British Museum Publications London}, 1981.


\bibitem{dessurt}
Davis, B., Morse, B., Price, B., Tensmeyer, C., Wigington, C., and Morariu, V.: End-to-end document recognition and understanding with Dessurt. In \textit{Proceedings of the European Conference on Computer Vision}, pages 280--296, 2022. Springer.


\bibitem{dan}
Coquenet, D., Chatelain, C., and Paquet, T.: Dan: A segmentation-free document attention network for handwritten document recognition. \textit{IEEE Transactions on Pattern Analysis and Machine Intelligence}, \textbf{45}(7), 8227--8243, 2023. IEEE.


\bibitem{trocr}
Li, M., Lv, T., Chen, J., Cui, L., Lu, Y., Florencio, D., Zhang, C., Li, Z., and Wei, F.: Trocr: Transformer-based optical character recognition with pre-trained models. In \textit{Proceedings of the AAAI Conference on Artificial Intelligence}, volume 37, number 11, pages 13094--13102, 2023.

\bibitem{got}
Wei, H., Liu, C., Chen, J., Wang, J., Kong, L., Xu, Y., Ge, Z., Zhao, L., Sun, J., Peng, Y., et al.: General OCR Theory: Towards OCR-2.0 via a Unified End-to-End Model, 2024.



\bibitem{gpt3}
Brown, T., Mann, B., Ryder, N., Subbiah, M., Kaplan, J.D., Dhariwal, P., Neelakantan, A., Shyam, P., Sastry, G., Askell, A., et al.: Language models are few-shot learners. Advances in Neural Information Processing Systems, \textbf{33}, 1877--1901 (2020)

\bibitem{icl_explanation}
Xie, Sang Michael and Raghunathan, Aditi and Liang, Percy and Ma, Tengyu : An explanation of in-context learning as implicit bayesian inference arXiv preprint arXiv:2111.02080 (2021)



\bibitem{flan}
Wei, J., Bosma, M., Zhao, V. Y., Guu, K., Yu, A. W., Lester, B., Du, N., Dai, A. M., and Le, Q. V.: Finetuned language models are zero-shot learners. \textit{arXiv preprint arXiv:2109.01652}, 2021.

\bibitem{opt}
Zhang, S., Roller, S., Goyal, N., et al.: OPT: Open pre-trained transformer language models. arXiv preprint \textbf{arXiv:2205.01068} (2022)

\bibitem{llama}
Touvron, H., Lavril, T., Izacard, G., Martinet, X., Lachaux, M.-A., Lacroix, T., Rozière, B., Goyal, N., Hambro, E., Azhar, F., et al.: Llama: Open and efficient foundation language models. \textit{arXiv preprint arXiv:2302.13971}, 2023.

\bibitem{llama2}
Touvron, H., Martin, L., Stone, K., Albert, P., Almahairi, A., Babaei, Y., Bashlykov, N., Batra, S., Bhargava, P., Bhosale, S., et al.: Llama 2: Open foundation and fine-tuned chat models. \textit{arXiv preprint arXiv:2307.09288}, 2023.




\bibitem{donut}
Kim, G., Hong, T., Yim, M., Nam, J., Park, J., Yim, J., Hwang, W., Yun, S., Han, D., and Park, S.: OCR-free document understanding transformer. In \textit{Proceedings of the European Conference on Computer Vision}, pages 498--517, 2022. Springer.


\bibitem{ureader}
Ye, J., Hu, A., Xu, H., Ye, Q., Yan, M., Xu, G., Li, C., Tian, J., Qian, Q., Zhang, J., et al.: Ureader: Universal OCR-free visually-situated language understanding with multimodal large language model. \textit{arXiv preprint arXiv:2310.05126}, 2023.

\bibitem{daniel}
Constum, T., Tranouez, P., and Paquet, T.: DANIEL: A fast Document Attention Network for Information Extraction and Labelling of handwritten documents. \textit{International Journal on Document Analysis and Recognition (IJDAR)}, pp. 1--23, 2025.


\bibitem{blip2}
Li, J., Li, D., Savarese, S., and Hoi, S. (2023d). Blip-2: Bootstrapping language-image pre-training with frozen image encoders and large language models. \textit{arXiv preprint arXiv:2301.12597}.

\bibitem{llava}
Li, C., Wong, C., Zhang, S., Usuyama, N., Liu, H., Yang, J., Naumann, T., Poon, H., and Gao, J.: Llava-med: Training a large language-and-vision model for biomedicine in one day. \textit{arXiv preprint arXiv:2306.00890}, 2023b.

\bibitem{e2str}
Zhao, Y., Wang, Q., Jin, L., et al.: Multi-modal In-Context Learning Makes an Ego-evolving Scene Text Recognizer. arXiv preprint \textbf{arXiv:2311.10363} (2023)


\bibitem{flamingo}
Alayrac, J.-B., Donahue, J., Luc, P., Miech, A., Barr, I., Hasson, Y., Leutenegger, S., Dieleman, S., Simonyan, K., Vinyals, O., et al.: Flamingo: a Visual Language Model for Few-Shot Learning. Advances in Neural Information Processing Systems, \textbf{35}, 23716--23730 (2022)

\bibitem{openflamingo}
Awadalla, A., Gao, I., Gardner, J., et al.: Openflamingo: An open-source framework for training large autoregressive vision-language models. arXiv preprint \textbf{arXiv:2308.01390} (2023)


\bibitem{otter}
Li, B., Zhang, Y., Chen, L., et al.: Mimic-it: Multi-modal in-context instruction tuning. arXiv preprint \textbf{arXiv:2306.05425} (2023)



\bibitem{idefics}
Laurençon, H., Saulnier, L., Tronchon, L., et al.: Obelics: An open web-scale filtered dataset of interleaved image-text documents. Advances in Neural Information Processing Systems, \textbf{36} (2024)

\bibitem{idefics2}
Lauren{\c{c}}on, H., Tronchon, L., Cord, M., and Sanh, V.: What matters when building vision-language models? \textit{arXiv preprint arXiv:2405.02246}, 2024.

\bibitem{kosmos}
Huang, S., Dong, L., Wang, W., et al.: Language is not all you need: Aligning perception with language models. Advances in Neural Information Processing Systems \textbf{36} (2024)


\bibitem{mmicl}
Zhao, H., Cai, Z., Si, S., et al.: Mmicl: Empowering vision-language model with multi-modal in-context learning. arXiv preprint \textbf{arXiv:2309.07915} (2023)


\bibitem{emu2}
Sun, Q., Cui, Y., Zhang, X., Zhang, F., Yu, Q., Wang, Y., Rao, Y., Liu, J., Huang, T., and Wang, X.: Generative Multimodal Models are In-Context Learners. \textit{Proceedings of the IEEE/CVF Conference on Computer Vision and Pattern Recognition (CVPR)}, pp. 14398-14409, June 2024.


\bibitem{mm1}
McKinzie, B., Gan, Z., Fauconnier, J.-P., Dodge, S., Zhang, B., Dufter, P., Shah, D., Du, X., Peng, F., Belyi, A., et al.: MM1: Methods, analysis, and insights from multimodal LLM pre-training. \textit{European Conference on Computer Vision}, pp. 304--323, 2024.

\bibitem{blip3}
Xue, L., Shu, M., Awadalla, A., Wang, J., Yan, A., Purushwalkam, S., Zhou, H., Prabhu, V., Dai, Y., Ryoo, M. S., et al.: XGen-MM (BLIP-3): A family of open large multimodal models. \textit{arXiv preprint arXiv:2408.08872}, 2024.


\bibitem{whatmakes}
Baldassini, A., Yuan, K., Raffel, C., Tenney, I.: What Makes Multimodal In-Context Learning Work? arXiv preprint \textbf{arXiv:2401.06538} (2024)

\bibitem{introspection}
Xu, W., Li, Y., Chen, A., Yuan, Z., Xu, J., et al.: From Introspection to Best Practices: Principled Analysis of Demonstrations in Multimodal In-Context Learning. arXiv preprint \textbf{arXiv:2402.01234} (2024)

\bibitem{link}
Tai, Yan and Fan, Weichen and Zhang, Zhao and Liu, Ziwei : Link-context learning for multimodal llms \textit{Proceedings of the IEEE/CVF Conference on Computer Vision and Pattern Recognition}pp. 27176--27185 (2024)


\bibitem{ocrvqa}
Mishra, A., Shekhar, S., Singh, A. K., Chakraborty, A.: OCR-VQA: Visual question answering by reading text in images. In \textit{2019 International Conference on Document Analysis and Recognition (ICDAR)}, pp. 947--952 (2019), \textbf{doi: 10.1109/ICDAR.2019.00156}

\bibitem{okvqa}
Schwenk, D., Khandelwal, A., Clark, C., Marino, K., Mottaghi, R.: A-okvqa: A benchmark for visual question answering using world knowledge. In \textit{Computer Vision -- ECCV 2022: 17th European Conference, Tel Aviv, Israel, October 23--27, 2022, Proceedings, Part VIII}, pp. 146--162, Springer-Verlag, Berlin, Heidelberg (2022), \textbf{doi: 10.1007/978-3-031-20074-8\_9}, \url{https://doi.org/10.1007/978-3-031-20074-8_9}


\bibitem{docvqa}
Mathew, M., Karatzas, D., Jawahar, C. V.: DocVQA: A dataset for VQA on document images. In \textit{2021 IEEE Winter Conference on Applications of Computer Vision (WACV)}, pp. 2199--2208 (2021), \textbf{doi: 10.1109/WACV48630.2021.00225}




\end{thebibliography}
%
\bibliographystyle{plain}     % Style classique (auteur-année non imposé)

\newpage
\section{\appendixname}

\subsection{Implementation Details} 
The Rosetta model, which incorporates a Context-Aware Tokenizer (CAT), a Visual Prompt Generator (VPG), and a Transformer Decoder, contains 240 million parameters. The model's vocabulary size is set by the CAT to 39, comprising 13 special tokens for data processing and 26 label tokens \(\langle t_{i} \rangle\), corresponding to the maximum number of unique symbols that can appear in a given context. In the VPG, images are divided into patches of $14 \times 14$. The vision encoder (ViT) is composed of 8 layers with an embedding size of 1280, which are initialized with pre-trained weights from \href{https://huggingface.co/Qwen/Qwen2-VL-2B-Instruct}{Qwen2-VL-2B-Instruct}, extracted from the original 32-layer model. Aside from the VPG, the entire model is trained from scratch.  The decoder consists of 8 Transformer decoder layers with an embedding size of 576, each comprising a causal self-attention mechanism, a feed-forward MLP, and layer normalization. The implementation of our model leverages code from \href{https://huggingface.co/Qwen/Qwen2-VL-2B-Instruct}{Qwen2-VL-2B-Instruct}. Training was conducted on an NVIDIA A100 GPU.We employed a batch size of 16 and a learning rate of 5e-6. The model was trained during 4000 step, where each step corresponding to 10k new query images \(X\) associated with 10k of context images \(X_c\).  
For optimization, we used the AdamW optimizer with a weight decay of 0.01, coupled with a cosine learning rate scheduler without warm-up. The OCR-based model contains a similar number of parameters as the Rosetta model and was trained in a comparable manner. For the training data, the font sizes were randomly selected between 20 and 30, with an average image resolution of 52 × 505 pixels.

\subsection{Quantitative Results of Rosetta on the Evaluation sets} 

\begin{figure}[h]
  \centering
  \resizebox{\textwidth}{!}{\includegraphics{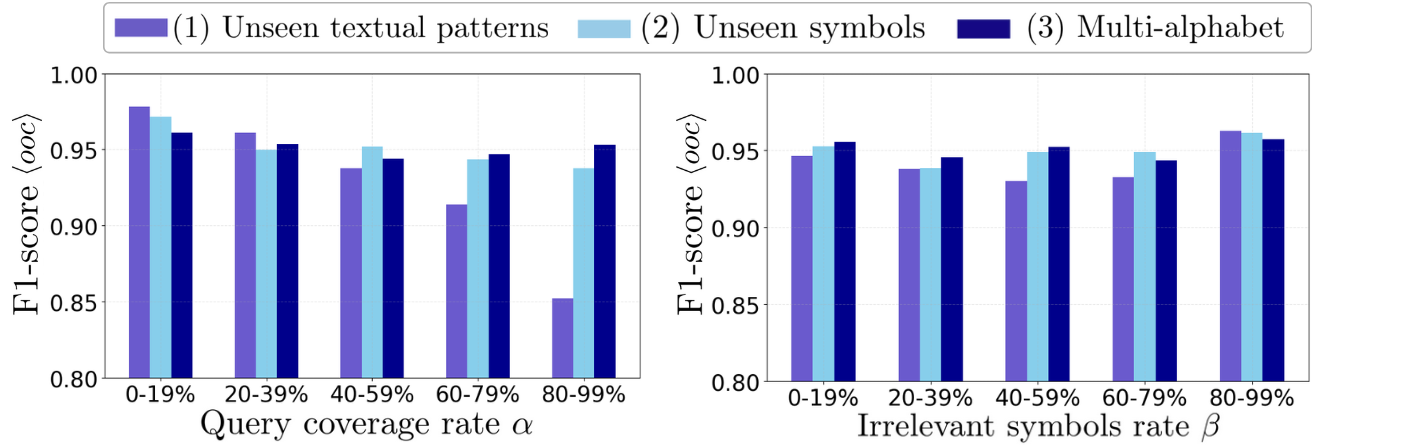}}
  \caption{F1-score on the \( \langle ooc \rangle \) token for the Rosetta model, evaluated on the three test sets under different context scenarios with varying query coverage rate  \(\alpha\)  and irrelevant symbol rate \(\beta\).}
  \label{fig:f1_score}
\end{figure}

\subsection{Qualitative Results of Rosetta on the Evaluation sets} 
\begin{figure}[h]
  \centering
  \resizebox{\textwidth}{!}{\includegraphics{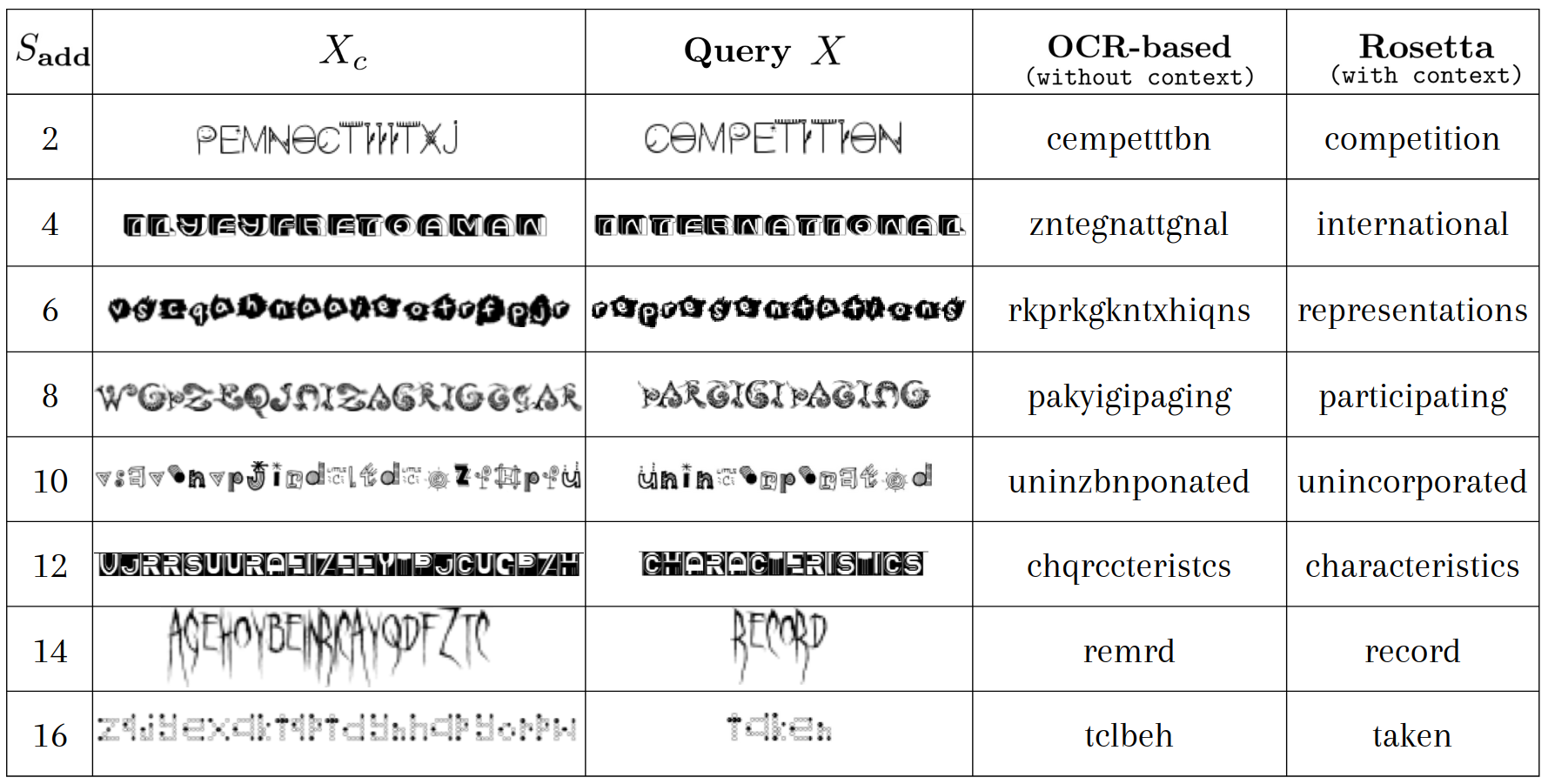}}
  \caption{Qualitative results on the textual test set between the OCR-based model train without context and the Rosetta model with a   query coverage rate of 100\%}
  \label{fig:vs}
\end{figure}
\vspace{-35mm}

\end{document}